\documentclass[sigconf]{acmart}
\AtBeginDocument{%
  }

\acmSubmissionID{6980}
\renewcommand\footnotetextcopyrightpermission[1]{}

\usepackage{pifont}
\usepackage{microtype}
\usepackage{graphicx}
\usepackage{amsmath}
\usepackage{amsfonts}
\usepackage{bbm}
\usepackage{booktabs}
\usepackage{multirow}
\usepackage{colortbl}
\usepackage{subcaption}
\usepackage{listings}
\usepackage{xcolor}
\usepackage{caption}
\graphicspath{{pics/}}
\numberwithin{equation}{section}

\usepackage{amsmath,amsfonts,bm}

\def\eqref#1{equation~\ref{#1}}

\def\1{\bm{1}}

\DeclareMathAlphabet{\mathsfit}{\encodingdefault}{\sfdefault}{m}{sl}
\SetMathAlphabet{\mathsfit}{bold}{\encodingdefault}{\sfdefault}{bx}{n}

\newcommand{\Tref}[1]{Tab.~\ref{#1}}

\newcommand{\Fref}[1]{Fig.~\ref{#1}}
\newcommand{\Sref}[1]{Sec.~\ref{#1}}

\newcommand{\eg}{\textit{e}.\textit{g}.}

\newcommand{\wrt}{\textit{w}.\textit{r}.\textit{t}\space}
\newcommand{\toolns}{\textsc{SafeGen}}
\newcommand{\tool}{\toolns\space}
\newcommand{\monens}{\textsc{Context Grounded End State Reasoning}}
\newcommand{\mone}{\monens\space}
\newcommand{\mtwons}{\textsc{End State Conditioned Video Evolution}}
\newcommand{\mtwo}{\mtwons\space}

\author{Jiangfan Liu}
\affiliation{%
  \institution{Beihang University, China}
  \city{}
  \country{}
}
\email{liujiangfan@buaa.edu.cn}
\orcid{0000-0002-1968-150X}

\author{Zexuan Cui}
\affiliation{%
  \institution{Beihang University, China}
  \city{}
  \country{}
}
\email{zxtsui@bjfu.edu.cn}
\orcid{0009-0001-3573-7657}

\author{Tianyuan Zhang}
\affiliation{%
  \institution{Beihang University, China}
  \city{}
  \country{}
}
\email{zhangtianyuan@buaa.edu.cn}
\orcid{0000-0001-9874-6828}

\author{Zonglei Jing}
\affiliation{%
  \institution{Beihang University, China}
  \city{}
  \country{}
}
\email{raykr@buaa.edu.cn}
\orcid{0009-0002-7487-5945}

\author{Zonghao Ying}
\affiliation{%
  \institution{Beihang University, China}
  \city{}
  \country{}
}
\email{yingzonghao@buaa.edu.cn}
\orcid{0009-0007-7393-7362}

\author{Yaoyuan Zhang}
\affiliation{%
  \institution{Beihang University, China}
  \city{}
  \country{}
}
\email{yaoyzhang@buaa.edu.cn}
\orcid{0009-0009-6014-1407}

\author{Jiakai Wang}
\affiliation{%
  \institution{Zhongguancun Laboratory, China}
  \city{}
  \country{}
}
\email{wangjk@mail.zgclab.edu.cn}
\orcid{0000-0001-5884-3412}

\author{Xiaoqi Jiang}
\affiliation{%
  \institution{Chery Automobile Co., Ltd., China}
  \city{}
  \country{}
}
\email{jiangxiaoqi1@mychery.com}
\orcid{0009-0000-4485-1075}

\author{Aishan Liu}
\authornote{The corresponding author.}
\affiliation{%
  \institution{Beihang University, China}
  \city{}
  \country{}
}
\email{liuaishan@buaa.edu.cn}
\orcid{0000-0002-4224-1318}

\author{Xianglong Liu}
\affiliation{%
  \institution{Beihang University, China}
  \city{}
  \country{}
}
\affiliation{%
  \institution{Zhongguancun Laboratory, China}
  \city{}
  \country{}
}
\email{xlliu@buaa.edu.cn}
\orcid{0000-0001-8425-4195}

\renewcommand{\shortauthors}{Jiangfan Liu et al.}

\begin{document}

\title{\toolns: Goal-Conditioned Video Diffusion of Safety-Critical Scenarios for VLM-Based Autonomous Driving}

\ccsdesc[500]{Security and privacy~Social aspects of security and privacy}
\ccsdesc[500]{Computing methodologies~Scene understanding}

\keywords{Safety-Critical Scenario Generation, Autonomous Driving}

\copyrightyear{2026}
\acmYear{2026}
\setcopyright{acmlicensed}\acmConference[MM '26]{Proceedings of the 34th ACM International Conference on Multimedia}{November 10-November 14, 2026}{Rio de Janeiro, Brazil}
\acmBooktitle{Proceedings of the 34th ACM International Conference on Multimedia (MM '26), November 10-November 14, 2026, Rio de Janeiro, Brazil}

\begin{abstract}
Vision-language models (VLMs) are increasingly deployed in autonomous driving (AD) systems, creating an urgent need for rigorous safety evaluation under rare yet safety-critical scenarios. Among these, interactions with vulnerable road users (VRUs), such as pedestrians, represent a major source of real-world failures. However, existing safety-critical scenario generation methods predominantly rely on simulator-based pipelines, which suffer from a substantial sim-to-real gap and often fail to capture realistic, diverse, and unforeseen human–vehicle interaction dynamics. To address this, we present \toolns, a goal-conditioned diffusion framework for safety-critical scenario generation in vision-language model-based autonomous driving systems (VLMADs). Our key insight is to formulate scenario generation as a goal-conditioned diffusion process, where a predefined catastrophic end-state (\eg, a collision with a pedestrian) serves as a strong supervisory signal, guiding the generation of temporally coherent video trajectories that naturally evolve toward safety-critical outcomes. Building on this formulation, we introduce \monens, which leverages VLMs to analyze benign driving contexts and infer latent vulnerabilities in human–vehicle interactions, producing structured end-state specifications that induce high-risk scenarios. Conditioned on these targets, we further propose \mtwons, which grounds semantic threats into physically plausible visual dynamics. Specifically, we instantiate high-risk agents (\eg, VRUs) within the scene via depth-aware geometric projection, followed by boundary-conditioned diffusion to generate intermediate frames with consistent motion patterns and temporal coherence. Extensive experiments across 3 VLMADs demonstrate that \tool increases the Judge Overall Score, a metric using a VLM judge to evaluate VLMADs' understanding and decision-making, by 24.25\% on average compared to SoTA baselines. Furthermore, fine-tuning a VLMAD with \toolns-generated data improves performance in real-world driving scenes by an average of 15.9\%. Our code is available at \url{https://github.com/JoFrc/SafeGen}.

\end{abstract}

\maketitle
\section{Introduction}

\begin{figure}[!t]
  \centering
  \includegraphics[width=\linewidth]{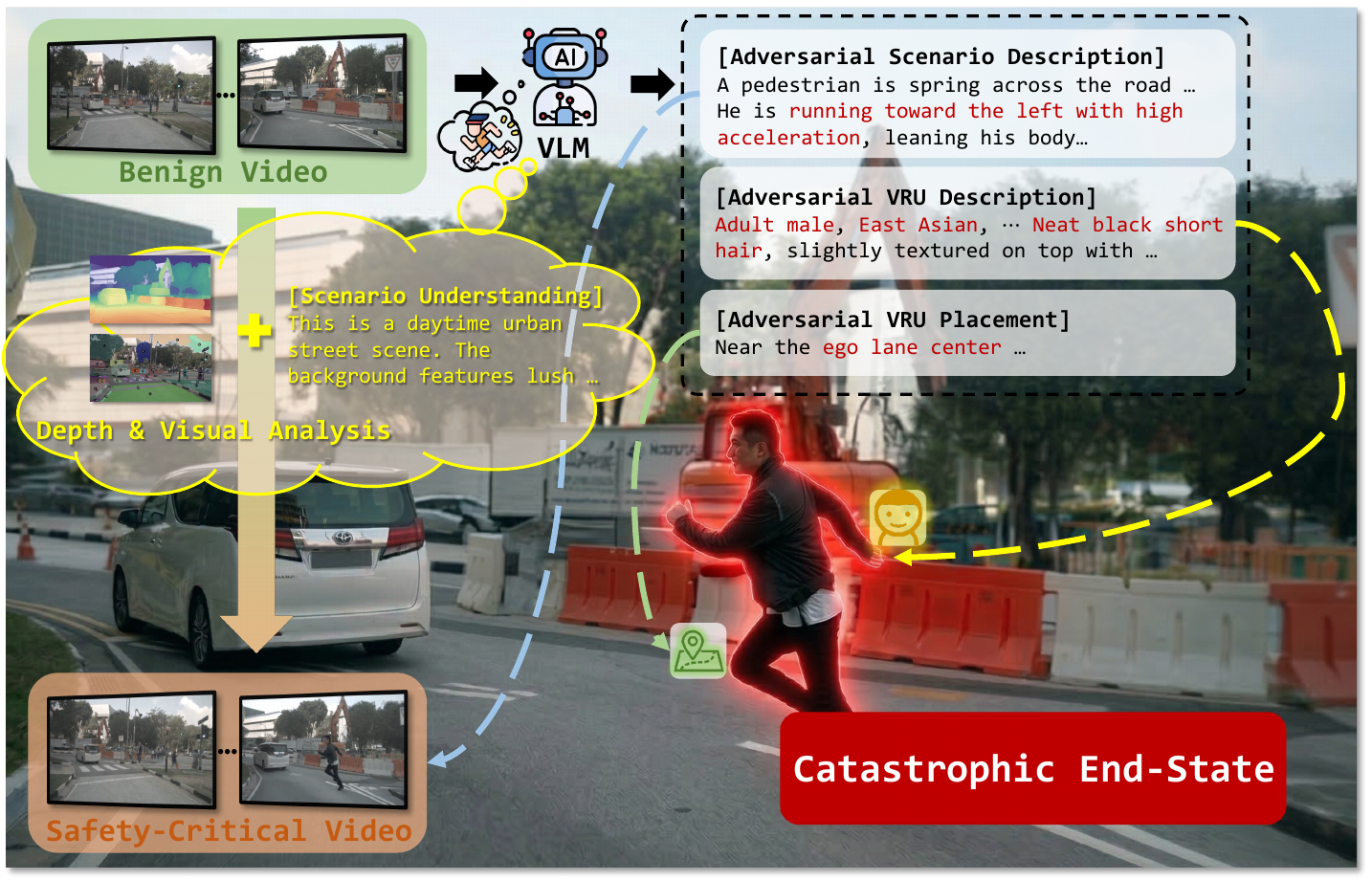}
  \Description{teaser image.}
  \caption{Given a benign driving video, \tool can synthesize a physically plausible and temporally coherent safety-critical scenario video for VLMAD evaluation.}
  \vspace{-0.1in}
  \label{fig: 1}
\end{figure}

As Vision-Language Models (VLMs) are widely deployed, their superior ability to perform sophisticated reasoning and decision-making in complex, open-world environments is rapidly advancing autonomous driving (AD) systems \cite{ma2024dolphins, ishaq2025drivelmm, gopalkrishnan2024multi}. These VLMADs leverage vast multimodal knowledge to interpret nuanced traffic semantics and navigate unpredictable urban scenarios. The deployment of these systems necessitates extensive evaluation of \emph{safety-critical scenarios} \cite{kalra2016driving, o2018scalable, ding2023survey}, particularly those involving human-vehicle interactions with Vulnerable Road Users (VRUs). Since VRUs exhibit highly nonlinear behaviors and complex intentions, generating diverse, high-fidelity safety-critical scenarios is not merely a testing requirement but a fundamental prerequisite for ensuring the reliability of VLMADs under high-risk, edge-case conditions.

However, existing approaches~\cite{chen2021adversarial, wang2021advsim, cao2022advdo, xiang2023v2xp, liu2026adversarial} predominantly rely on simulator-based pipelines, which struggle with a substantial sim-to-real gap, hindering their effectiveness. The synthetic textures and simplified physics fail to capture real-world visual complexity, while agents constrained by predefined maneuver templates and rigid simulator kinematics cannot replicate the unpredictable human-vehicle interactions. Consequently, such constrained environments yield evaluation data that neither reflects the realism nor captures unforeseen human-vehicle conflicts.

To address these, we propose \toolns, a goal-conditioned diffusion framework that formulates safety-critical scenario video synthesis as a boundary-conditioned state-evolution problem. By constraining generation between a benign start state and a specified adversarial end state, the framework models synthesis as a goal-conditioned diffusion generation problem. The framework first proposes \ding{182} \mone that identifies latent vulnerabilities by modeling spatiotemporal dependencies in dynamic traffic scenes. At this stage, the VLM performs high-level semantic reasoning over traffic context and identifies high-risk factors, such as occlusion zones. The specific frame containing the extracted factor is then defined as the critical last frame. The VLM then constructs an adversarial specification \wrt this last frame that maximizes collision risk while maintaining scene-level plausibility and logical coherence. Based on this, we introduce \ding{183} \mtwons, which translates abstract semantic threats into high-fidelity visual evidence through physically grounded generation. Conditioned on the adversarial end state as a geometric anchor, \tool synthesizes a realistic VRU and integrates it into the scene via depth-aware projection and iterative latent inpainting, ensuring consistency with 3D geometry and ambient illumination. Finally, using an asymmetric conditioning strategy, \tool performs boundary-conditioned diffusion generation under boundary constraints to enforce strict kinematic alignment, yielding a synthesized video that smoothly evolves from a nominal driving state to the specified catastrophic outcome.

Extensive experiments across 3 representative VLMADs demonstrate the superior efficacy of \toolns, which increases the Judge Overall Score by 24.25\% on average compared to SoTA baselines, exposing critical vulnerabilities in the reasoning and planning modules of VLMADs. Additionally, human validation of the synthesized videos indicates comparably high threat and visual fidelity, demonstrating that these machine-generated risks pose fundamental safety challenges for experienced drivers. Finally, fine-tuning a VLMAD on \toolns-generated data improves performance by 15.9\% in real-world scenes, effectively augmenting the training distribution with high-impact corner cases that are essential for survival in unforeseen environments. Our \textbf{contributions} are:

\begin{itemize}
  \item We introduce \toolns, a goal-conditioned diffusion framework that generates safety-critical scenario videos by reformulating scenario generation as an evolutionary process.

  \item We propose a pipeline that combines {\mone} for vulnerability discovery with {\mtwo} for kinematically consistent, temporally coherent video generation.

  \item Extensive evaluations show that \tool effectively exposes safety flaws, increasing the JOS by 24.25\% on average compared to baseline methods.
\end{itemize}

\section{Related Works}

\subsection{Generative Video Synthesis}

Recent video diffusion models can forecast driving scenes by generating future frames conditioned on an initial state. Researchers refer to this setting as First-Frame-to-Video (F2V)~\cite{brooks2024video}, in which the model uses only the first frame and samples a plausible continuation of the scene. Although F2V is effective for open-ended prediction, it offers limited control over terminal outcomes. Consequently, reliably synthesizing rare safety-critical events (\eg, specific collisions) remains difficult, as stochastic sampling may produce different end states. First-Last-Frame-to-Video (FLF2V) extends this formulation by conditioning generation on both the first and last frames~\cite {wan2025wan, zhang2025packing}. In our terminology, both F2V and FLF2V are boundary-conditioned variants with 1 and 2 boundary anchors, respectively. In FLF2V, intermediate frames are generated via boundary-conditioned diffusion sampling, yielding temporally coherent videos that satisfy explicit boundary constraints.

\subsection{Safety-Critical Scenario Generation}

Traditional safety-critical scenario generation is largely simulator-based, in which challenging corner cases are mined by controlling agents and varying environmental parameters. Existing pipelines typically employ optimization-based search to escalate threats, evolving across diverse techniques: reinforcement learning~\cite{ding2020learning}, gradient-free methods~\cite{wang2021advsim}, and gradient-based trajectory optimization~\cite{zhang2022adversarial}, and recently, approaches combining LLM-driven semantic reasoning with multi-agent collaborative optimization~\cite{liu2026adversarial}. While this paradigm provides strong controllability and effectively stress-tests specific modules, it inherently suffers from a sim-to-real gap rooted in simulator~\cite{Dosovitskiy17, li2023metadrive} limitations such as synthetic textures, simplified sensor artifacts, and constrained maneuver templates. Consequently, these simulator-based methods struggle to fully capture the context-dependent visual complexities of real-world traffic. Meanwhile, recent generative world models~\cite{wang2024drivedreamer, gao2023magicdrive} prioritize visual realism and general scene synthesis over explicit safety-critical threat escalation, leaving a gap in evaluating driving models against emergent, high-fidelity hazards. Therefore, they are not directly aligned with the core problem addressed in this work.

\textbf{Comparison.} Our framework differs from prior works in three key aspects: \ding{182} \textbf{Motivation.} Simulator-based methods mainly aim to stress-test specific modules under predefined maneuvers, whereas \tool targets high-level semantic reasoning vulnerabilities in VLMADs by synthesizing unexpected human-vehicle interactions. \ding{183} \textbf{Technical implementation.} Instead of relying on parametric action spaces, \tool enables adversarial intent to evolve naturally. \ding{184} \textbf{Performance.} By photorealistically injecting adversarial VRUs into real-world benign videos, \tool produces more diverse and realistic failure modes.

\section{Methodology}

\begin{figure*}[h]
  \centering
  \includegraphics[width=\linewidth]{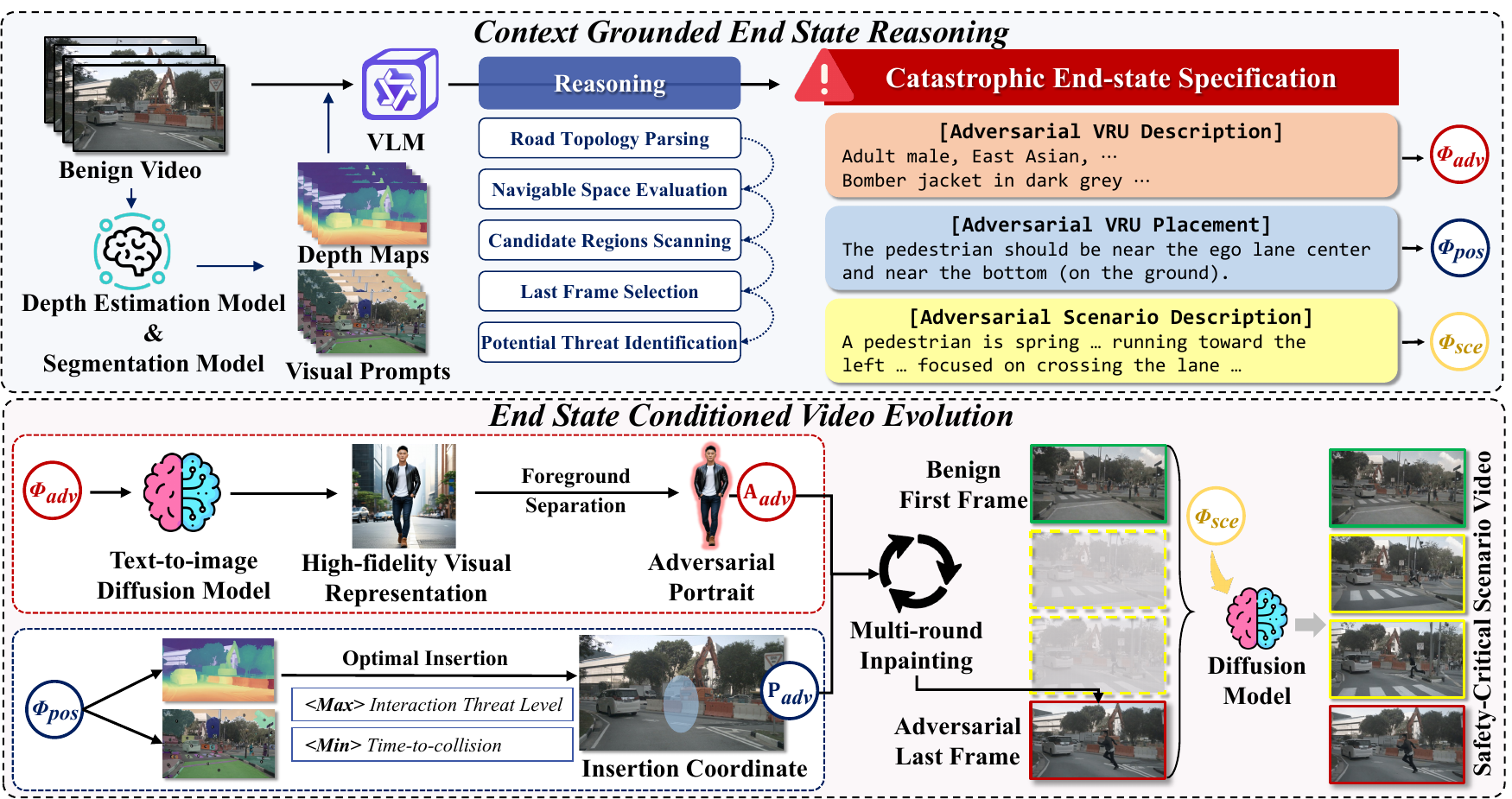}
  \Description{framework}
  \caption{Overview of the proposed \tool framework for safety-critical scenario video generation.}
  \label{fig: framework}
\end{figure*}

\subsection{Problem Definition}

Let $\mathbf{s}=\{{\Phi}_{\text{sce}}, {\Phi}_{\text{adv}}, {\Phi}_{\text{pos}} \mid \hat{T}\}$ denote the \textbf{catastrophic end-state specification}, derived from a benign driving video $\mathcal{V}=\{\mathbf{I}_t\}_{t=0}^{T}$. This end state specification $\mathbf{s}$ comprises the description of \ding{182} the evolution process of a safety-critical scenario ${\Phi}_{\text{sce}}$, \ding{183} a VRU portrait ${\Phi}_{\text{adv}}$, and \ding{184} the insertion region ${\Phi}_{\text{pos}}$. These elements are all conditioned on a selected critical last frame $\mathbf{I}_{\hat{T}}$, where $\hat{T} \in (0, T]$. To pinpoint this critical last frame $\mathbf{I}_{\hat{T}}$, a VLM $f_{vlm}$ analyzes the benign driving video sequence $\mathcal{V}$ to determine where a catastrophic end-state is most probable. This semantic reasoning is physically grounded by spatial and instance-level priors from a depth estimation model $f_{dep}$ and a segmentation model $f_{seg}$.

Building upon the established specification $\mathbf{s}$, \tool synthesizes the safety-critical video sequence $\hat{\mathcal{V}}=\{\hat{\mathbf{I}}_t\}_{t=0}^{\hat{T}}$, preserving the first frame as $\hat{\mathbf{I}}_0=\mathbf{I}_0$. Specifically, a diffusion model $f_{sd}$ initially generates an adversarial portrait $\mathbf{A}_{adv}$ conditioned on ${\Phi}_{\text{adv}}$, which is subsequently scaled utilizing the camera intrinsic matrix $\mathbf{K}$. To ensure physical plausibility, we constrain our spatial search to the intersection of the designated target region ${\Phi}_{\text{pos}}$ and valid planar ground surfaces $\mathcal{M}_{\text{ground}}$. Within this localized area, we determine the optimal insertion coordinate $\mathbf{P}_{adv}$ by evaluating the estimated depth information and the ego-vehicle velocity $v_{ego}$ to maximize the interaction threat level. The visual portrait $\mathbf{A}_{adv}$ and the spatial coordinate $\mathbf{P}_{adv}$ are then seamlessly integrated to construct the highly realistic adversarial last frame $\hat{\mathbf{I}}_{\hat{T}}$. Finally, guided by the scenario description ${\Phi}_{\text{sce}}$, a boundary-conditioned video diffusion model $f_{vd}$ interpolates the intermediate dynamics. This model generates a temporally coherent sequence that smoothly evolves from the benign first frame $\mathbf{I}_{0}$ toward the synthesized adversarial last frame $\hat{\mathbf{I}}_{\hat{T}}$, yielding the complete safety-critical scenario video $\hat{\mathcal{V}}$. The overview of our proposed \tool is illustrated in \Fref{fig: framework}.

\subsection{Context Grounded End State Reasoning}

Given a benign driving video $\mathcal{V}=\{\mathbf{I}_t\}_{t=0}^{T}$, \tool first constructs the scenario context by extracting both physical and semantic priors. To achieve this, we employ a depth estimation model $f_{dep}$ and a segmentation model $f_{seg}$ to process $\mathcal{V}$, yielding depth maps $\{f_{dep}(\mathbf{I}_t)\}_{t=0}^{T}$ and visual prompts $\{f_{seg}(\mathbf{I}_t)\}_{t=0}^{T}$. These spatial and semantic priors facilitate the transition from 2D pixel observations to 3D physical understanding, significantly enhancing the reasoning capabilities of standard VLMs. Leveraging this enriched context, \mone employs a VLM $f_{vlm}$ to reason potential threats to human-vehicle interaction by analyzing the video sequence to identify a specific critical frame $\mathbf{I}_{\hat{T}}$, where a catastrophic end state $\mathbf{s}$ is highly probable. Guided by the depth and segmentation priors, $f_{vlm}$ effectively parses the road topology, evaluates navigable spaces, and systematically scans for candidate regions prone to interaction conflicts. This ensures the identified threats are physically plausible and highly threatening, ultimately yielding the catastrophic end-state specification:
\begin{equation}
\begin{aligned}
    \mathbf{s} &= \{{\Phi}_{\text{sce}}, {\Phi}_{\text{adv}}, {\Phi}_{\text{pos}} \mid \hat{T}\}\\
    &=f_{vlm}(\mathcal{V}, \{f_{dep}(\mathbf{I}_t)\}_{t=0}^{T}, \{f_{seg}(\mathbf{I}_t)\}_{t=0}^{T}).
\end{aligned}
\end{equation}

This process involves the detailed instantiation of 3 core components within the reasoned catastrophic end-state specification $\mathbf{s}$ at the selected last frame $\mathbf{I}_{\hat{T}}$. The safety-critical scenario description ${\Phi}_{\text{sce}}$ describes the evolution of the threat from a benign initial state to the defined catastrophic end state, explicitly specifying the adversarial agent's intent, motion patterns, and collision dynamics. The description of the VRU portrait ${\Phi}_{\text{adv}}$ defines the visual and semantic attributes of the adversarial entity, detailing nuanced characteristics required to synthesize a photorealistic and contextually appropriate threat. Finally, the target region ${\Phi}_{\text{pos}}$ designates a localized candidate search area, defines the specific high-risk spatial bounds within the scene, explicitly narrowing the search space for the subsequent optimal insertion of the adversarial actor.

Unlike standard forward prediction, which typically defaults to safe, high-probability trajectories, our approach explicitly infers latent vulnerabilities. Rather than projecting uncertain futures from current conditions, we deterministically anchor a predefined catastrophic end-state at the critical frame $\mathbf{I}_{\hat{T}}$. Serving as a strong supervisory signal, this anchor enables the framework to reason backward and produce a structured end-state specification $\mathbf{s}$ that induces high-risk scenarios. Ultimately, \mone transforms abstract vulnerabilities into explicit spatial and semantic constraints, providing a blueprint for grounding semantic threats into physically plausible, temporally coherent video that naturally evolves toward safety-critical outcomes.

\subsection{End State Conditioned Video Evolution}

Building on the established catastrophic end-state specification $\mathbf{s}$, \tool utilizes a text-to-image (T2I) diffusion model $f_{sd}$ to synthesize an initial adversarial portrait conditioned on the VRU description ${\Phi}_{\text{adv}}$. This generative phase transforms abstract semantic attributes into a high-fidelity visual representation. Crucially, the diffusion model ensures that the synthesized entity possesses realistic textures, appropriate structural poses, and visual characteristics that strictly align with the intended collision dynamics dictated by the overarching scenario ${\Phi}_{\text{sce}}$. To facilitate seamless physical integration into the target 3D environment, we subsequently perform precise foreground separation on the generated output. This isolation process systematically removes irrelevant background artifacts, ultimately yielding the refined, background-free adversarial portrait $\mathbf{A}_{adv}$ prepared for downstream spatial alignment.

To determine the exact optimal insertion coordinate $\mathbf{P}_{adv}$, we seamlessly bridge the semantic reasoning with physical constraints. We initially leverage the localized candidate search area ${\Phi}_{\text{pos}}$. Within this specific spatial bound, we evaluate the corresponding depth map $f_{dep}(\mathbf{I}_{\hat{T}})$ to identify planar regions exhibiting low depth variance. This defines a spatial constraint set, $\mathcal{M}_{\text{ground}}$, that represents valid, drivable ground surfaces. To maximize the interaction threat level, we finalize $\mathbf{P}_{adv}$ by selecting the exact spatial coordinate within this intersection that minimizes the time-to-collision (TTC). This spatial optimization process is formulated as:
\begin{equation}
    \mathbf{P}_{adv} = \mathop{\arg\min}_{\mathbf{P} \in {\Phi}_{\text{pos}} \cap \mathcal{M}_{\text{ground}}} \frac{f_{dep}(\mathbf{I}_{\hat{T}})(\mathbf{P})}{v_{ego}},
\end{equation}
where $f_{dep}(\mathbf{I}_{\hat{T}})(\mathbf{P})$ extracts the localized depth value at the spatial coordinate $\mathbf{P}$, and $v_{ego}$ represents the ego-vehicle velocity. Utilizing the background-free portrait $\mathbf{A}_{adv}$ and the spatial geometry derived from the camera intrinsic matrix $\mathbf{K}$, we explicitly project the adversarial entity onto the optimized point $\mathbf{P}_{adv}$. Subsequently, we execute a multi-round inpainting process on the selected critical frame $\mathbf{I}_{\hat{T}}$. This iterative operation ensures physical consistency in ambient lighting, shadows, and color distribution, ultimately constructing the highly realistic adversarial last frame $\hat{\mathbf{I}}_{\hat{T}}$.

With the temporal boundaries firmly established by the benign first frame $\mathbf{I}_0$ and the synthesized adversarial last frame $\hat{\mathbf{I}}_{\hat{T}}$, we employ a boundary-conditioned video diffusion model $f_{vd}$ to interpolate the dynamic progression of the adversarial interaction. Guided by the overarching safety-critical scenario description ${\Phi}_{\text{sce}}$, this goal-coditioned generation is formalized as:
\begin{equation}
    \begin{aligned} 
    \hat{\mathcal{V}}&=\{\hat{\mathbf{I}}_t\}_{t=0}^{\hat{T}}\\ 
    &=f_{vd}(\mathbf{I}_0, \hat{\mathbf{I}}_{\hat{T}} \mid {\Phi}_{\text{sce}}).
    \end{aligned}
\end{equation}
The video diffusion model processes these structural boundaries and semantic conditions to generate intermediate frames that form the complete safety-critical scenario video $\hat{\mathcal{V}}$. This evolutionary generation maintains strict temporal consistency and physically plausible human-vehicle interactions throughout the entire sequence.

\subsection{Systemic Generation Workflow}

The systemic generation workflow of \tool transforms a benign driving video $\mathcal{V}$ into a safety-critical scenario video $\hat{\mathcal{V}}$ by first extracting physical and semantic priors using a depth estimation model $f_{dep}$ and a segmentation model $f_{seg}$. Provided with this enriched context, a VLM $f_{vlm}$ reasons under a reverse simulation paradigm to identify a critical frame $\mathbf{I}_{\hat{T}}$ containing potential threats, which is then assigned to the last frame of the adversarial safety-critical scenario video. It then synthesizes the catastrophic end-state specification $\mathbf{s}$, which explicitly defines the overarching scenario description ${\Phi}_{\text{sce}}$, the visual attributes of the adversarial agent ${\Phi}_{\text{adv}}$, and the localized candidate search region ${\Phi}_{\text{pos}}$.

To instantiate this threat, a text-to-image diffusion model $f_{sd}$ generates a background-free adversarial portrait $\mathbf{A}_{adv}$ conditioned on ${\Phi}_{\text{adv}}$. Concurrently, the exact insertion coordinate $\mathbf{P}_{adv}$ is localized by minimizing the TTC within the intersection of the search region ${\Phi}_{\text{pos}}$ and valid ground surfaces $\mathcal{M}_{\text{ground}}$. The portrait is geometrically scaled via the camera intrinsic matrix $\mathbf{K}$, projected onto $\mathbf{P}_{adv}$, and integrated into $\mathbf{I}_{\hat{T}}$ through multi-round inpainting to construct the highly realistic adversarial last frame $\hat{\mathbf{I}}_{\hat{T}}$. Finally, utilizing the benign first frame $\mathbf{I}_0$ and $\hat{\mathbf{I}}_{\hat{T}}$ as structural anchors, a boundary-conditioned video diffusion model $f_{vd}$ interpolates the intermediate dynamic progression of the adversarial human-vehicle interaction. Guided by ${\Phi}_{\text{sce}}$, $f_{vd}$ generates a kinematically plausible trajectory that smoothly evolves the scene toward the targeted collision state, outputting the final sequence $\hat{\mathcal{V}}$.
\section{Experiments}

\begin{figure*}[ht]
  \centering
  \includegraphics[width=\linewidth]{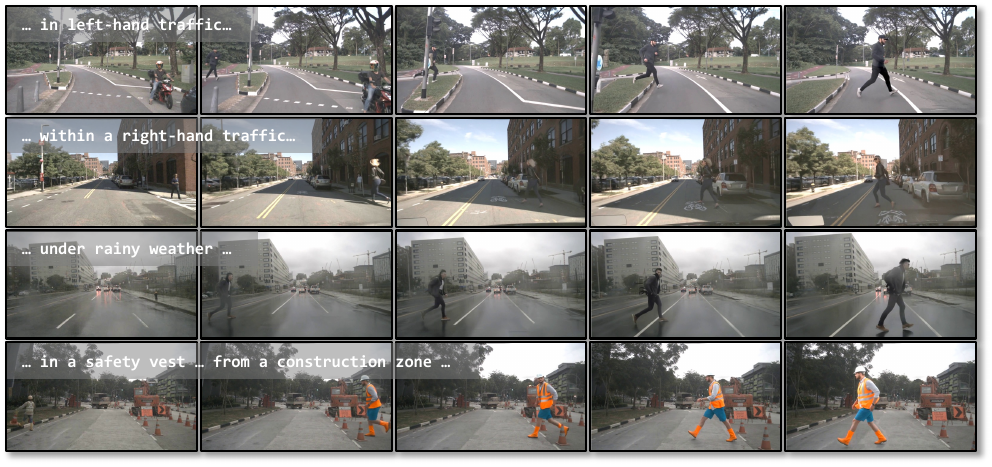}
  \Description{framework}
  \caption{Safety-critical scenario videos generated by \tool across construction, rain, and different traffic orientations.}
  \label{fig: visualization}
\end{figure*}

\subsection{Experimental Setup}
\label{sec: Experimental Setup}

\quad \textbf{Implementation details.} We initially use DepthAnythingV3~\cite{lin2025depth} and MobileSAM~\cite{zhang2023faster} to provide depth maps and visual prompts, respectively. Building upon this spatial context, we employ Qwen3-VL-32B~\cite{Qwen3VL} for semantic threat identification. To physically materialize these identified threats, pedestrian assets are generated using Stable Diffusion 3.5~\cite{esser2024scaling} and anchored in the scene via multi-round inpainting. Finally, Wan2.1-FLF2V~\cite{wan2025wan} bridges the entire sequence via boundary-conditioned diffusion. We generate 850 safety-critical scenarios from the nuScenes dataset~\cite{caesar2020nuscenes}, with each video standardized to a resolution of $1280 \times 720$ at 10 FPS (81 frames in total).

\textbf{VLMAD algorithms.} We evaluate \tool using 3 representative reasoning-centric VLMADs based on the question-answering (QA) paradigm, chosen to cover diverse architectural designs and reasoning mechanisms. \textbf{Dolphins}~\cite{ma2024dolphins} is a conversational framework that processes video tokens and text instructions for grounded reasoning. \textbf{DriveLMM-o1}~\cite{ishaq2025drivelmm} formulates AD as a graph-based QA task to establish causal relationships between perception and planning. \textbf{EM-VLM4AD}~\cite{gopalkrishnan2024multi} adopts a LLaVA-based instruction-following approach to map visual observations into language-based control tokens. These models are selected to assess the impact of deterministic safety-critical conflicts on their logical consistency.

\textbf{Compared baselines.} We evaluate \tool against 4 baselines: \textbf{ Learning-to-Collide (LC)}~\cite{ding2020learning}, \textbf{AdvSim (AS)}~\cite{wang2021advsim}, \textbf{Adversarial Trajectory Optimization (AT)}~\cite{zhang2022adversarial}, and \textbf{ScenGE}~\cite{liu2026adversarial}. To ensure fair comparisons, we generate 850 scenarios per baseline using an open-loop setup that replays original ego trajectories to produce the final threat videos. Furthermore, we adapt these predominantly vehicle-to-vehicle baselines to conflicts centered around human-vehicle interactions.


\textbf{Evaluation Metrics.} Evaluating safety-critical generative videos presents a unique dilemma: traditional simulator testing relies on closed-loop physical collision metrics, rendering it infeasible for open-loop videos, whereas generative methods~\cite{wang2024drivedreamer, gao2023magicdrive} primarily prioritize visual fidelity over safety-critical logic. To bridge this gap, we propose the Judge Overall Score (JOS~\textcolor{red}{$\uparrow$}) in \Sref{sec: Main Results and Analysis}. JOS employs a high-capacity VLM as an automated judge to assess the logical consistency between visual threats and the agent's textual reasoning. It systematically evaluates Perception Alignment (PA~\textcolor{blue}{$\downarrow$}), Anticipation Error (AE~\textcolor{red}{$\uparrow$}), and Planning Compliance (PC~\textcolor{blue}{$\downarrow$}) (see Appendix for details). Formulated as $\text{JOS} = 0.2(10-\text{PA}) + 0.3\text{AE} + 0.5(10-\text{PC})$, this metric heavily penalizes critical planning violations to specifically expose decision-making vulnerabilities within the textual responses of QA-based VLMADs. Comprehensive ablation studies on different hyperparameter settings are provided in the supplementary material. To ensure visual fidelity and physical-scenario threat, we follow standard experimental protocols to evaluate TTC~\textcolor{blue}{$\downarrow$}, DRAC~\textcolor{red}{$\uparrow$}, FID~\textcolor{blue}{$\downarrow$}, and FVD~\textcolor{blue}{$\downarrow$}, as detailed in \Sref{sec: Scenario Quality and Diversity}. \textbf{Higher \textcolor{red}{$\uparrow$} indicates worse VLMADs performance, while \textcolor{blue}{$\downarrow$} indicates the contrary.}

\subsection{Main Results and Analysis}
\label{sec: Main Results and Analysis}

\begin{table*}[ht]
  \centering
  \caption{Main results on the adversarial efficacy for different VLMADs. We report the Judge Overall Score (JOS) along with its sub-metrics. We also report the change in performance ($\Delta$) relative to the baseline. The \tool columns are shaded in gray. Improvements and degradations are highlighted in red and blue, respectively.}
  \label{tab: main_results}
  \resizebox{\textwidth}{!}{
    \begin{tabular}{c ccccc ccccc ccccc}
      \toprule[1.2pt]
      & \multicolumn{5}{c}{Dolphins} & \multicolumn{5}{c}{DriveLMM-o1} & \multicolumn{5}{c}{EM-VLM4AD} \\
      \cmidrule(lr){2-6} \cmidrule(lr){7-11} \cmidrule(lr){12-16}
      \multirow{-2}{*}{Metric} & LC & AS & AT & ScenGE & \cellcolor{gray!15}\textbf{\toolns} & LC & AS & AT & ScenGE & \cellcolor{gray!15}\textbf{\toolns} & LC & AS & AT & ScenGE & \cellcolor{gray!15}\textbf{\toolns} \\
      \midrule
      \textbf{JOS}~$\textcolor{red}\uparrow$ & 6.26 & 6.45 & 7.59 & 7.51 & \cellcolor{gray!15}\textbf{9.03} & 5.05 & 4.96 & 5.00 & 7.41 & \cellcolor{gray!15}\textbf{7.96} & 8.00 & 9.05 & 9.04 & 9.32 & \cellcolor{gray!15}\textbf{9.61} \\
      $\Delta$ & / & +0.19 & +1.33 & +1.25 & \cellcolor{gray!15}{\color[HTML]{700000} \textbf{+2.77}} & / & -0.09 & -0.05 & +2.36 & \cellcolor{gray!15}{\color[HTML]{700000} \textbf{+2.91}} & / & +1.05 & +1.04 & +1.32 & \cellcolor{gray!15}{\color[HTML]{700000} \textbf{+1.61}} \\
      \midrule
      PA~$\textcolor{blue}\downarrow$ & 5.20 & 4.82 & 2.91 & 2.57 & \cellcolor{gray!15}\textbf{1.26} & 6.81 & 6.93 & 6.92 & 2.99 & \cellcolor{gray!15}\textbf{2.39} & 3.85 & 1.58 & 1.52 & 0.93 & \cellcolor{gray!15}\textbf{0.42} \\
      $\Delta$ & / & -0.38 & -2.29 & -2.63 & \cellcolor{gray!15}{\color[HTML]{000070} \textbf{-3.94}} & / & +0.12 & +0.11 & -3.82 & \cellcolor{gray!15}{\color[HTML]{000070} \textbf{-4.42}} & / & -2.27 & -2.33 & -2.92 & \cellcolor{gray!15}{\color[HTML]{000070} \textbf{-3.43}} \\
      \midrule
      AE~$\textcolor{red}\uparrow$ & 7.08 & 7.12 & 8.04 & 5.77 & \cellcolor{gray!15}\textbf{9.08} & 7.43 & 7.26 & 7.16 & 7.68 & \cellcolor{gray!15}\textbf{7.97} & 8.35 & 9.28 & 9.22 & 9.08 & \cellcolor{gray!15}\textbf{9.62} \\
      $\Delta$ & / & +0.04 & +0.96 & -1.31 & \cellcolor{gray!15}{\color[HTML]{700000} \textbf{+2.00}} & / & -0.17 & -0.27 & +0.25 & \cellcolor{gray!15}{\color[HTML]{700000} \textbf{+0.54}} & / & +0.93 & +0.87 & +0.73 & \cellcolor{gray!15}{\color[HTML]{700000} \textbf{+1.27}} \\
      \midrule
      PC~$\textcolor{blue}\downarrow$ & 3.65 & 3.45 & 2.48 & 1.41 & \cellcolor{gray!15}\textbf{0.88} & 5.64 & 5.66 & 5.53 & 2.59 & \cellcolor{gray!15}\textbf{1.91} & 1.48 & 0.83 & 0.84 & 0.44 & \cellcolor{gray!15}\textbf{0.39} \\
      $\Delta$ & / & -0.20 & -1.17 & -2.24 & \cellcolor{gray!15}{\color[HTML]{000070} \textbf{-2.77}} & / & +0.02 & -0.11 & -3.05 & \cellcolor{gray!15}{\color[HTML]{000070} \textbf{-3.73}} & / & -0.65 & -0.64 & -1.04 & \cellcolor{gray!15}{\color[HTML]{000070} \textbf{-1.09}} \\
      \bottomrule[1.2pt]
    \end{tabular}
  }
\end{table*}

As shown in \Tref{tab: main_results}, \tool consistently achieves the highest Judge Overall Score (JOS~\textcolor{red}{$\uparrow$}) across all evaluated VLMADs, yielding a \textbf{24.25\% improvement on average} compared to the baseline methods. Since the JOS metric strictly penalizes misalignments between textual reasoning and visual threats, these high scores expose a systemic failure rather than a superficial drop in accuracy. Specifically, \tool drives the JOS to 9.03 on Dolphins, 7.96 on DriveLMM-o1, and 9.61 on EM-VLM4AD, completely outperforming the strongest baseline in each respective group. This consistent vulnerability reveals that the identified reasoning flaws are inherent to current multimodal paradigms. Consequently, our goal-conditioned scenarios go beyond triggering isolated perception errors. Rather, \tool actively induces severe, cascading reasoning failures throughout the entire decision-making pipeline.

\Fref{fig: visualization} provides critical qualitative evidence explaining this adversarial efficacy. The generated adversarial pedestrians maintain strong semantic alignment with their surroundings. They naturally adjust their clothing, placement, and motion to blend into complex environments, including construction zones, rainy weather, and varying traffic orientations. Since these intrusions are visually plausible and physically grounded, they effortlessly bypass the initial anomaly detection of VLMADs. Consequently, models misinterpret these agents as benign, failing to anticipate their trajectories until a collision is kinematically unavoidable.

\textbf{Perception.} Perception Alignment (PA~$\textcolor{blue}\downarrow$) measures the capacity of a VLMAD to identify and spatially localize critical objects within the driving scene. The experimental results demonstrate that \tool severely degrades this perception capability. The proposed framework \textbf{reduces the PA score by 65.38\% on average} relative to the baseline average. DriveLMM-o1 exhibits the most significant absolute perception failure, experiencing a \textbf{drop from a baseline average score of 5.91 to 2.39}. This performance degradation occurs because the generated pedestrians maintain geometric grounding and blend photorealistically into real-world driving frames. The depth-aware anchoring mechanism ensures the adversarial actor integrates seamlessly into the traffic topology, preventing the perception modules from flagging synthetic texture anomalies. As a result, agents fail to recognize the intrusion as a high-priority threat until the conflict becomes imminent.

\textbf{Prediction.} The Anticipation Error (AE~$\textcolor{red}\uparrow$) quantifies the deviation between the predicted intent of other actors and the actual adversarial trajectory. \tool maximizes AE, yielding an \textbf{average increase of 14.13\%} over the baseline average across all tested models. The Dolphins model demonstrates extreme vulnerability in this stage, where \tool triggers a \textbf{jump from a baseline average of 7.00 to 9.08}. The \mone module drives this increase by identifying latent vulnerabilities to plan trajectories that maximize collision likelihood. The framework generates intent that remains logically consistent with the traffic scene yet kinematically aggressive, forcing the VLM to misinterpret the impending threat. The reasoning logic fails because it cannot anticipate the goal-oriented evolution of the pedestrian path.

\textbf{Planning.} Planning Compliance (PC~$\textcolor{blue}\downarrow$) evaluates the safety of the VLMADs' emergency maneuvers when facing an imminent collision. Our findings confirm that \tool substantially diminishes this compliance under safety-critical stress. \tool \textbf{reduces the PC score by 62.59\% on average} relative to the baseline. DriveLMM-o1 exhibits a substantial absolute reduction in its compliance score, \textbf{from a baseline average of 4.86 to 1.91}. This adversarial efficacy stems directly from the deterministic target-setting of the generated scenario. The boundary-conditioned diffusion generation operates under semantic constraints to continuously steer the synthesized motion toward the intended adversarial end-state. This process preserves temporal and physical plausibility while severely degrading the planning capacity of VLMADs. The severely reduced compliance values demonstrate that the discovered vulnerabilities reflect a fundamental inability to generate safe responses under precisely controlled physical threats.

\subsection{Scenario Quality and Diversity}
\label{sec: Scenario Quality and Diversity}

While \Sref{sec: Main Results and Analysis} evaluates the reasoning vulnerabilities of VLMADs via the JOS metric, it is equally crucial to validate the intrinsic quality of the generated scenarios independent of the target driving models. Evaluating generative safety-critical data presents a unique challenge, as it requires bridging two distinct domains. Traditional AD metrics (\eg, TTC) rigorously quantify kinematic threats but cannot assess multi-agent semantic context. Conversely, standard video generation metrics (\eg, FID) prioritize pixel-level visual fidelity while completely ignoring the physical validity of the depicted threats. Since both evaluation paradigms fall short in isolation, establishing a comprehensive benchmark is necessary. To bridge this gap, we evaluate \tool from three orthogonal perspectives: kinematic urgency, visual realism, and semantic diversity. These model-agnostic evaluations verify that our synthesized videos are not merely visually convincing but also rigorously grounded, high-fidelity representations of physical hazards.

\begin{figure}
  \centering
  \begin{subfigure}{0.49\linewidth}
    \includegraphics[width=\linewidth]{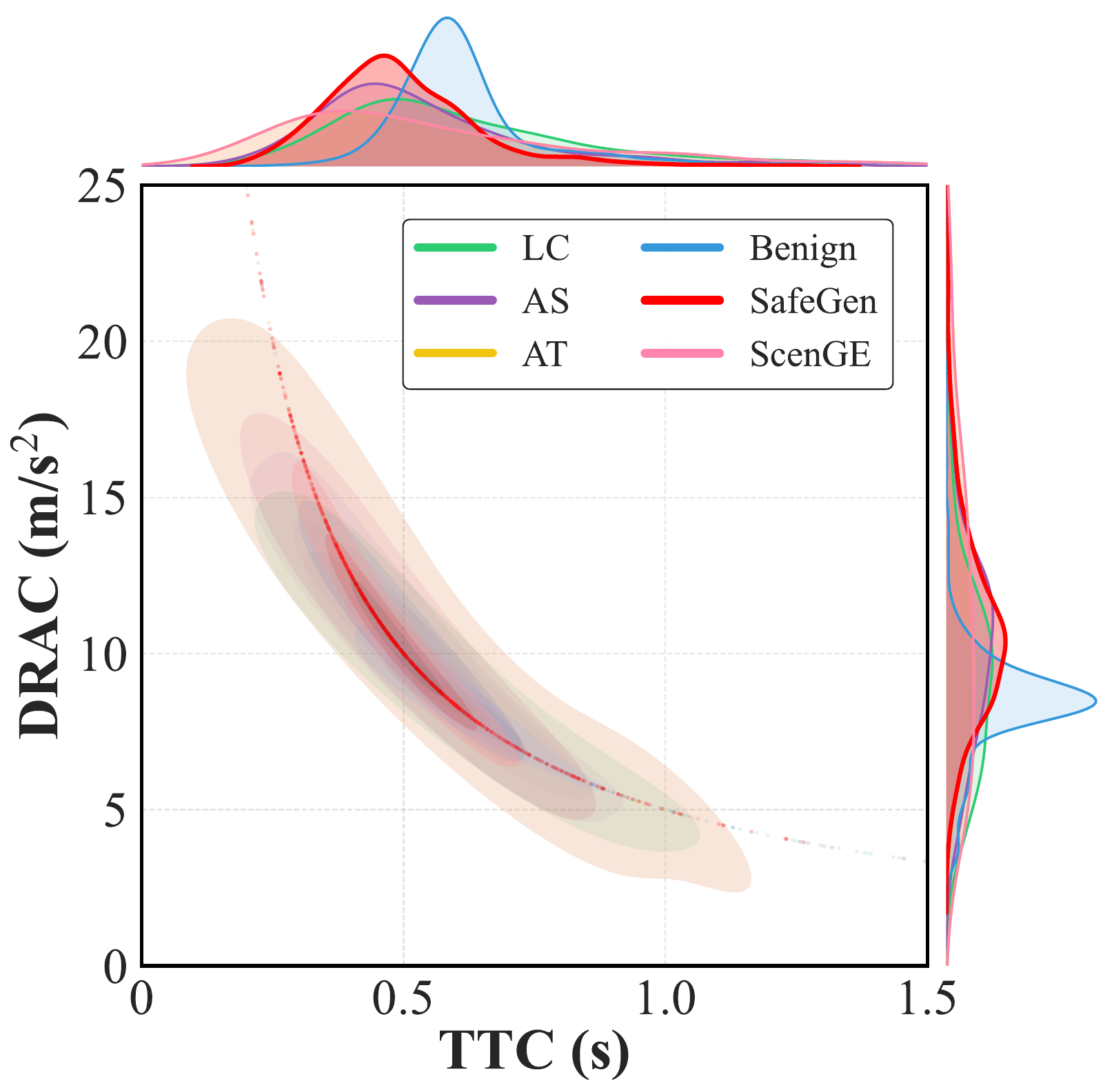}
    \Description{ttc-drac.}
    \caption{Comparison of TTC and DRAC statistics.}
    \label{fig: ttc-drac}
  \end{subfigure}
  \hfill
  \begin{subfigure}{0.49\linewidth}
    \includegraphics[width=\linewidth]{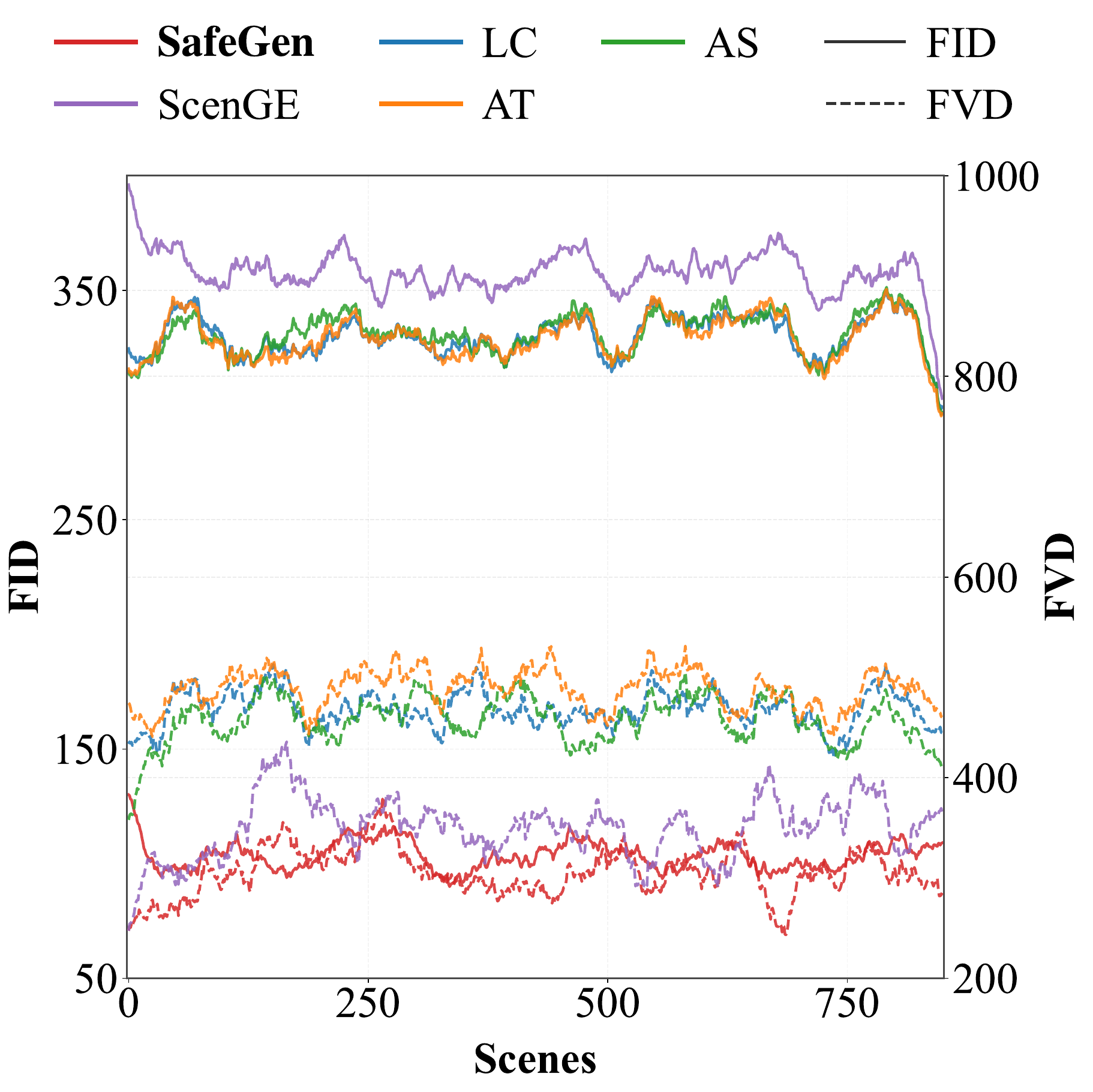}
    \Description{fid-fvd.}
    \caption{Comparison of FID and FVD statistics.}
    \label{fig: fid-fvd}
  \end{subfigure}
  
  \begin{subfigure}{\linewidth}
    \includegraphics[width=\linewidth]{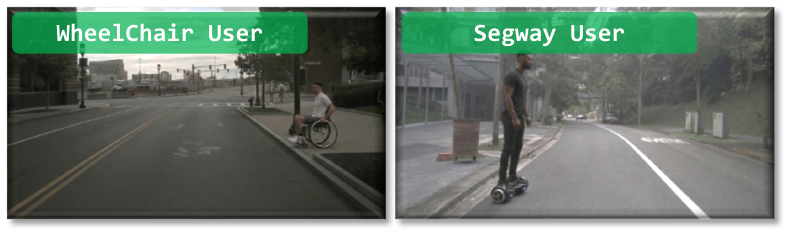}
    \Description{case-diversity.}
    \caption{Representative examples of VRU categories.}
    \label{fig: case-diversity}
  \end{subfigure}
  
  \Description{Threat and Realism Analysis.}
  \caption{Threat and realism analysis including kinematic metrics, visual metrics, and qualitative diversity examples.}
  \label{fig: threat-realism}
\end{figure}

\textbf{Kinematic Urgency Coverage.} A valid safety-critical scenario must unequivocally force an evasive maneuver. We quantify this physical threat level using Time-to-Collision (TTC~$\textcolor{blue}\downarrow$) and Deceleration Rate to Avoid Collision (DRAC~\textcolor{red}{$\uparrow$}). As illustrated in \Fref{fig: ttc-drac}, \tool induces a severe and deliberate distributional shift compared to the source data. The original benign driving logs reside in a low-risk domain, with TTC values densely clustered between 2.5 and 3.0s, and DRAC peaking well below 10 $\mathrm{m/s^2}$. In stark contrast, \toolns-generated scenarios drastically \textbf{compress the TTC into a highly critical window below 0.5s}, while simultaneously driving the required evasive \textbf{deceleration to extreme levels of 15-20 $\mathrm{m/s^2}$}. This pronounced quantitative escalation confirms that our framework effectively transforms safe, nominal driving trajectories into imminent collision events, entirely eliminating the reaction margins that VLMADs typically rely on. 

\textbf{Visual Fidelity.} High physical urgency is undermined if the generated threat lacks visual realism, as VLMADs might fail due to superficial pixel-level artifacts rather than genuine reasoning flaws. We measure visual quality using Fr\'echet Inception Distance (FID~$\textcolor{blue}\downarrow$) and Fr\'echet Video Distance (FVD~$\textcolor{blue}\downarrow$) across our complete dataset of 850 generated scenes (\Fref{fig: fid-fvd}). \tool demonstrates remarkable generation stability, achieving a \textbf{mean FID near 100 and a mean FVD of approximately 400}. More importantly, the distribution of these scores reveals a tight, consistent clustering across the entire dataset, avoiding the severe degradation often observed in unconstrained video synthesis. This persistent high fidelity validates our structural design. By anchoring the generation process with depth-aware spatial projections and iterative latent inpainting, \tool effectively suppresses boundary artifacts and temporal flickering, ensuring that the synthesized VRUs blend seamlessly and photorealistically into highly complex ambient environments.

\textbf{Generation Diversity.} Beyond physical metrics, the open-world robustness of a VLMAD heavily depends on its exposure to semantically diverse threats. Standard simulator-based pipelines are constrained by finite, manually rigged 3D asset libraries, resulting in repetitive and predictable hazard profiles. As demonstrated in \Fref{fig: case-diversity}, \tool overcomes this by synthesizing a rich spectrum of VRUs that naturally inhabit the long tail of driving distributions. By seamlessly instantiating high-fidelity, unconventional interaction scenarios, such as individuals on Segways, wheelchair users, or construction workers, \tool forces VLMADs to reason about novel semantic categories rather than simply memorizing standard pedestrian bounding boxes. This open-vocabulary generation capability ensures that our safety evaluation reflects the true, unpredictable diversity of real-world urban environments.

\subsection{Ablation Studies}

\begin{figure}
  \centering
  \includegraphics[width=\linewidth]{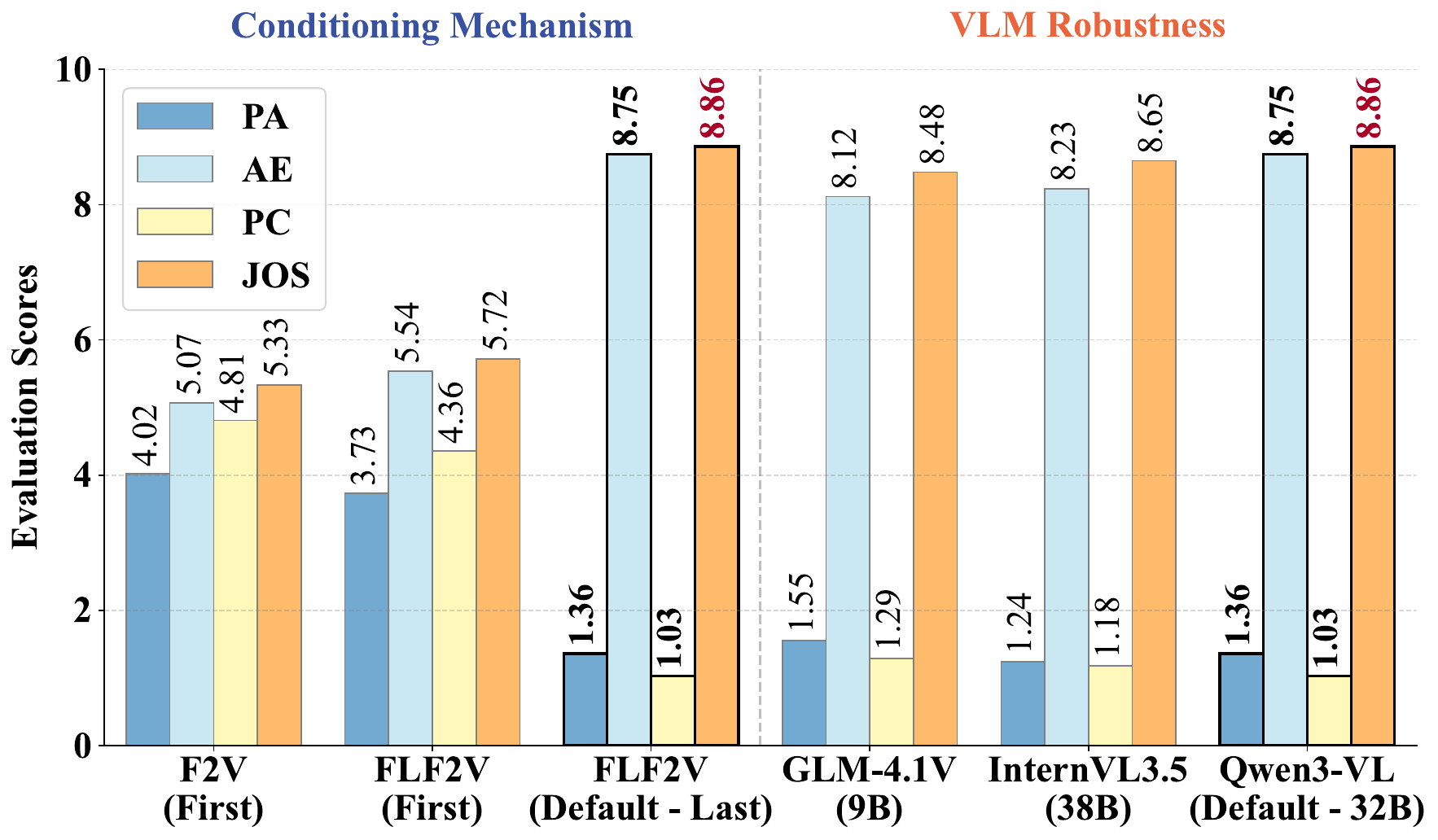}
  \Description{Ablation Studies.}
  \caption{Impact of generation conditioning mechanisms (left) and VLM choice (right) on the JOS and its sub-metrics.}
  \label{fig: ablation_studys}
\end{figure}

We conduct ablation studies to evaluate the necessity of the core mechanisms within \tool and to assess its sensitivity to specific component selections, as shown in \Fref{fig: ablation_studys}. To provide a comprehensive evaluation, we analyze the JOS and its detailed sub-metrics to uncover how the vulnerabilities manifest across the perception, prediction, and planning stages. We use the default \tool settings for all experiments unless otherwise specified.

\ding{182} \textbf{Mechanism of Boundary Conditioning.} We validate the goal-conditioned formulation by comparing the default \tool architecture against two alternative synthesis strategies. The first alternative is a First-Frame-to-Video (F2V) baseline initialized with an adversarial first frame. The second is a boundary-conditioned FLF2V baseline that injects the threat at the first frame and targets a benign last frame. \Fref{fig: ablation_studys} demonstrates that anchoring the generation with an adversarial last frame significantly outperforms both alternatives. Beyond yielding an \textbf{average increase of 60.36\%} in the JOS metric, the breakdown of sub-metrics reveals exactly how the threat materializes. The default FLF2V (Last) setup severely degrades VLMAD's perception and planning, reducing PA to 1.36 and PC to 1.03, while simultaneously increasing the anticipation error (AE) to 8.75. In contrast, the F2V configuration lacks a terminal constraint, and the FLF2V (First) baseline forces the model to interpolate toward a benign ending. Consequently, the absence of a deterministic end-state constraint leads to highly variable generation outcomes, resulting in a high frequency of non-collision events where the VLMAD can still maintain moderate PA and PC scores. Conversely, anchoring the catastrophic event strictly at the terminal state compels the intermediate visual dynamics to escalate toward a collision naturally. This result confirms the essential need for end-state conditioning in safety-critical synthesis.

\ding{183} \textbf{Robustness of VLM Reasoning.} We evaluate the dependence of the semantic threat reasoning module on specific VLMs. We replace the default Qwen3-VL-32B model with a smaller GLM-4.1V-9B-Thinking~\cite{hong2025glm} and an InternVL3.5-38B~\cite{wang2025internvl3_5}. \Fref{fig: ablation_studys} demonstrates that the overall JOS remains consistently high across these different architectures. Furthermore, the sub-metrics (PA, AE, PC) remain tightly clustered, indicating that this specific failure mode consistently degrades the anticipation and planning capacities across different reasoning engines. The alternative models produce no significant decrease in adversarial efficacy. This result shows that our generation method is independent of the selected VLM. The framework successfully identifies latent threats and plans effective catastrophic end states without relying on specific model structures or unique capability levels.

\subsection{Downstream Utility and Human Validation}

\begin{table}
  \centering
  \caption{Dolphins accuracy on VRU-Accident before and after fine-tuning on \toolns-generated data. Lower $\mathrm{ACC}_{i}$ values indicate worse VLMAD performance.}
  \label{tab: vru_accident}
  \resizebox{\columnwidth}{!}{
    \begin{tabular}{lcccc}
        \toprule
        Method & $\mathrm{ACC}_{\mathrm{AC}}$~\textcolor{blue}{$\downarrow$} & $\mathrm{ACC}_{\mathrm{PM}}$~\textcolor{blue}{$\downarrow$} & $\mathrm{ACC}_{\mathrm{AT}}$~\textcolor{blue}{$\downarrow$} & $\mathrm{ACC}_{\mathrm{AVG.}}$~\textcolor{blue}{$\downarrow$} \\
        \midrule
        Dolphins (Vanilla) & 29.1 & 43.6 & 43.8 & 38.8 \\
        Dolphins (\toolns) & \textbf{46.7} & \textbf{55.5} & \textbf{61.8} & \textbf{54.7} \\
        \bottomrule
    \end{tabular}
  }
\end{table}

\begin{figure}
  \centering
  \includegraphics[width=\linewidth]{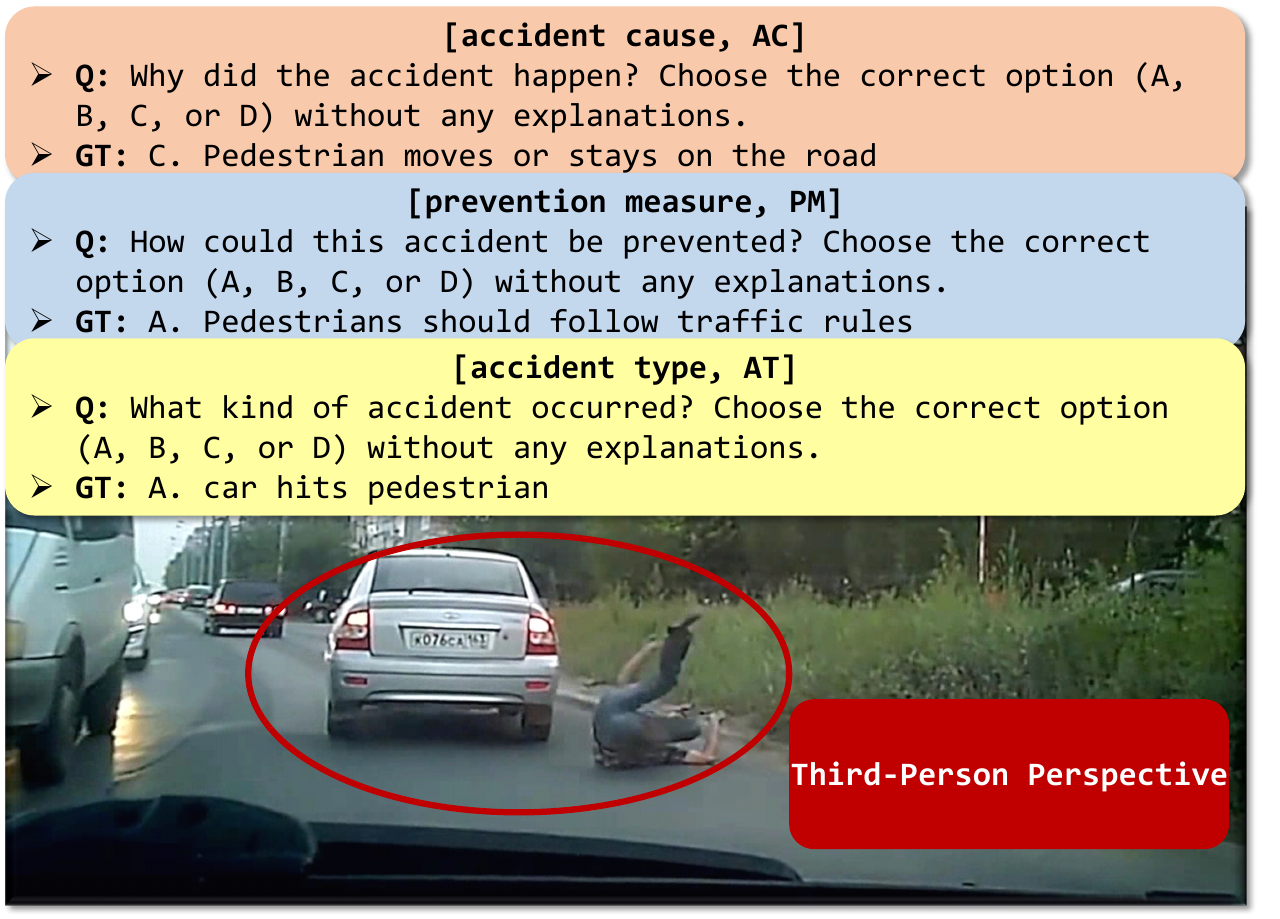}
  \Description{vru_example.}
  \caption{Example frame from VRU-Accident, a dataset containing mixed viewpoints such as third-person collisions, illustrating the 3 most relevant QA categories selected.}
  \label{fig: vru_example}
\end{figure}

\quad \textbf{Downstream Fine-Tuning.} To assess the downstream utility of \toolns-generated scenarios for model enhancement, we conduct a fine-tuning experiment using the Dolphins. We use the JOS Judge to automatically annotate high-quality QA training pairs. Dolphins is then fine-tuned on this synthesized dataset and evaluated on the unseen VRU-Accident benchmark~\cite{Kim_2025_ICCV}. To focus strictly on high-risk interactions, we select the 3 most safety-critical VQA categories: Accident Cause ($\mathrm{ACC}_{\mathrm{AC}}$~\textcolor{blue}{$\downarrow$}), Prevention Measure ($\mathrm{ACC}_{\mathrm{PM}}$~\textcolor{blue}{$\downarrow$}), and Accident Type ($\mathrm{ACC}_{\mathrm{AT}}$~\textcolor{blue}{$\downarrow$}). Notably, while \tool generates first-person ego-vehicle conflicts, the unseen VRU-Accident benchmark features a diverse mixture of camera viewpoints, including a portion of third-person human-vehicle collisions, as shown in \Fref{fig: vru_example}. Although this partial viewpoint mismatch constrains our absolute accuracy ceilings, \Tref{tab: vru_accident} shows a \textbf{15.9\% improvement in $\mathrm{ACC}_{\mathrm{AVG.}}$~\textcolor{blue}{$\downarrow$}}. This improvement confirms that \toolns-generated data transfers safety reasoning to unseen real-world domains, demonstrating its value for downstream model enhancement.

\begin{figure}
  \centering

  \begin{subfigure}{0.49\linewidth}
    \includegraphics[width=\linewidth]{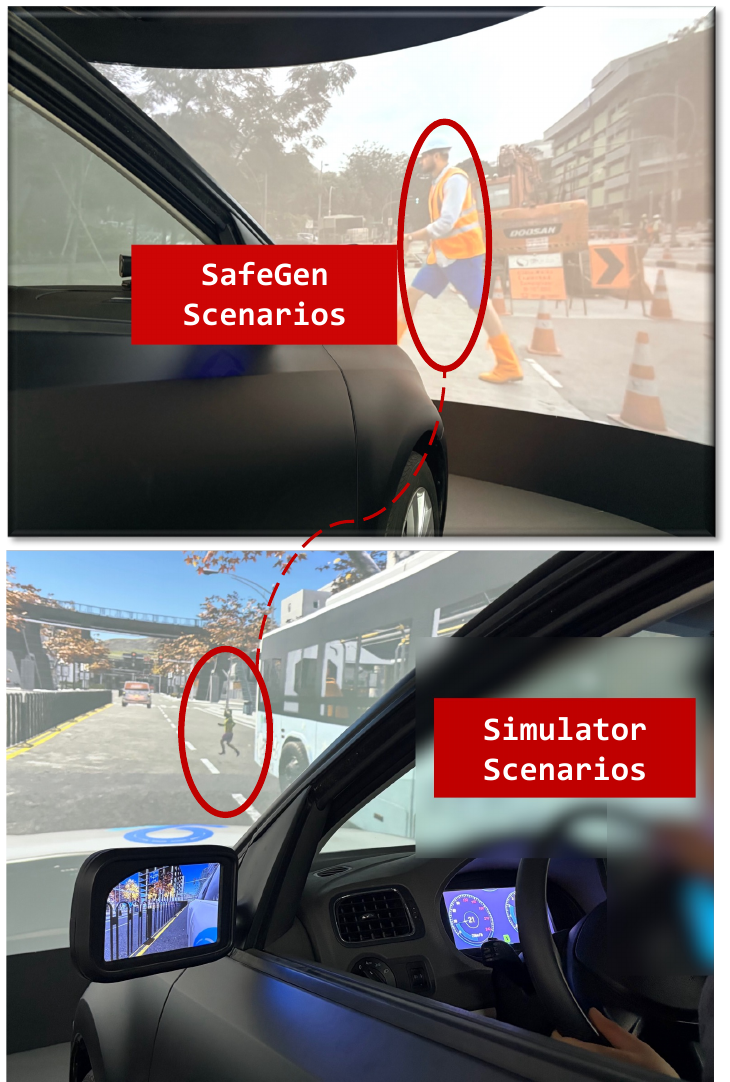}
    \Description{hitl.}
    \caption{Human-In-The-Loop platform with a surround display.}
    \label{fig: hitl}
  \end{subfigure}
  \hfill
  \begin{subfigure}{0.49\linewidth}
    \includegraphics[width=\linewidth]{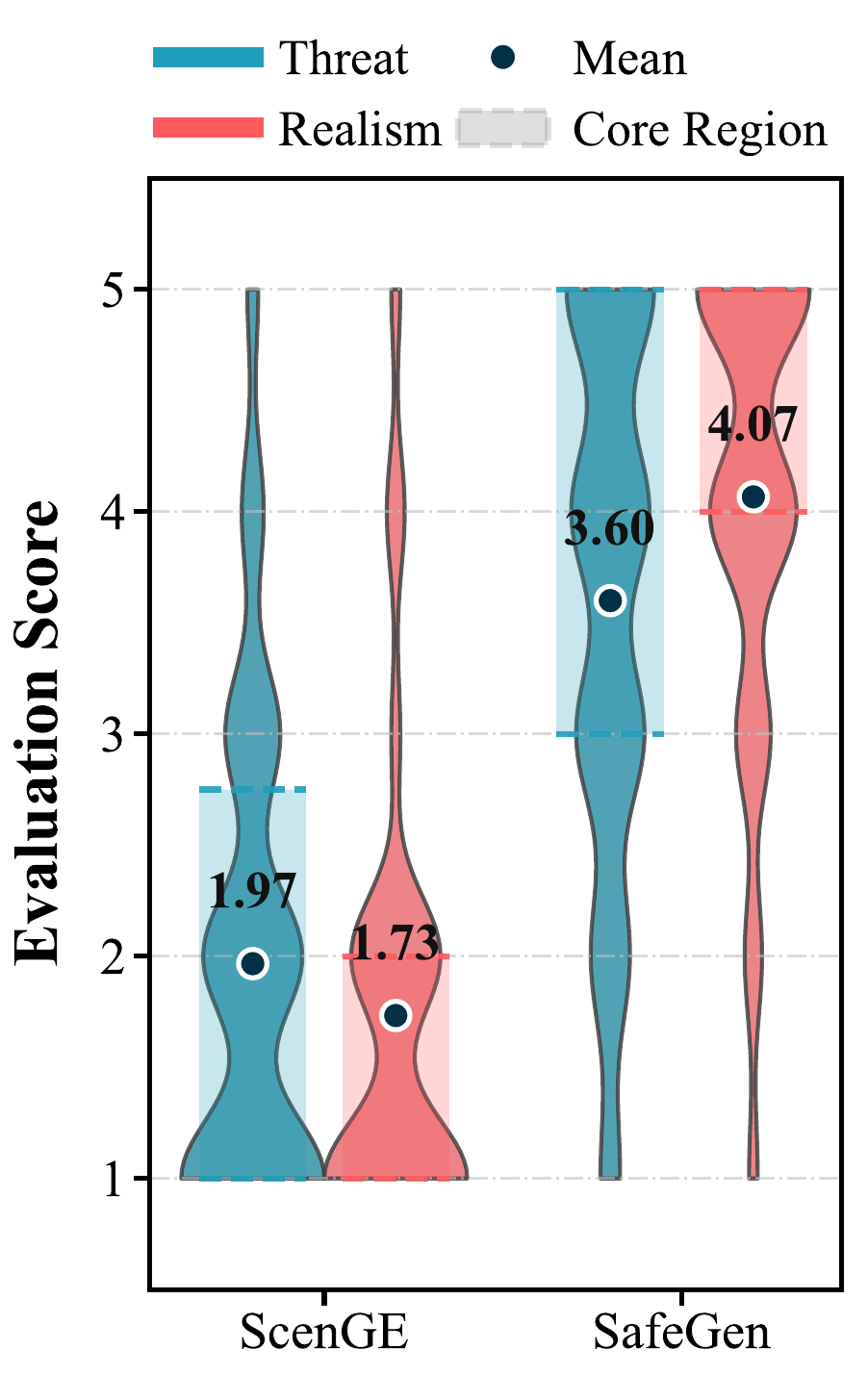}
    \Description{violin.}
    \caption{Plots of participant ratings on perceived threat and realism.}
    \label{fig: violin}
  \end{subfigure}
  \Description{Human Evaluation.}
  \caption{Human evaluation, with simulator setup on the left and rating distributions on the right.}
\end{figure}

\textbf{Human Evaluation.} While kinematic metrics provide objective measurements of physical urgency, establishing true scenario realism and threat validity requires human cognitive assessment. We conduct a human-in-the-loop (HITL) study involving 30 licensed drivers to compare the perceptual quality of \toolns-generated scenarios with the ScenGE baseline. To accurately replicate the cognitive load of real-world driving, the study utilizes a 210-degree immersive surround-display simulator equipped with an in-cabin control interface (\Fref{fig: hitl}). To mitigate anticipation bias and prevent participants from simply bracing for an inevitable crash, each driver is exposed to a continuous sequence containing 10 adversarial scenarios seamlessly mixed with 40 benign driving segments, with no prior knowledge of event timing. After each session, participants rate the perceived realism and threat intensity on a 5-point Likert scale. \Fref{fig: violin} illustrates the resulting score distributions, revealing a decisive advantage for our framework. Across both evaluation dimensions, the distribution of \tool is consistently concentrated at higher values than that of ScenGE. Qualitative feedback from participants highlighted that ScenGE scenarios often exhibited rigid pedestrian kinematics and unnatural spatiotemporal progression, thereby weakening the perceived urgency of the conflict. In stark contrast, \tool scenarios were consistently judged to be behaviorally plausible yet deeply threatening. The fluid, context-aware evolution of the generated VRUs successfully induced genuine surprise and strong human-perceived safety pressure, validating that our generative pipeline effectively bridges the sim-to-real gap in safety-critical evaluation.

\section{Conclusion}

We introduced \toolns, a diffusion-based framework for generating realistic safety-critical videos. By uncovering long-tail VRU failures, \tool provides more rigorous safety evaluations and improves VLMAD performance through fine-tuning. \textbf{Future work} includes extending the framework to multi-agent interactions.

\textbf{Ethical Statement.} We aim to improve VLMAD safety by identifying pre-deployment risks. We promote responsible, human-supervised use of our generation tools to ensure they are used solely for robustness enhancement.

\bibliographystyle{ACM-Reference-Format}
\bibliography{refs}

\end{document}